\documentclass[letterpaper, 10 pt, conference]{ieeeconf}

\IEEEoverridecommandlockouts

\usepackage{graphicx}
\usepackage{tikz}

\newcommand{\myvspace}{-0.1in}

\usepackage{amsmath,amssymb,mathabx,amsfonts}

\usepackage{cite}
\usepackage{hyperref}
\usepackage[capitalise]{cleveref}

\usepackage{booktabs}
\usepackage{multirow}

\usepackage{caption}
\usepackage{acronym}

\acrodef{pddl}[PDDL]{Planning Domain Definition Language}
\acrodef{mdp}[MDP]{Markov Decision Process}
\acrodef{ged}[GED]{Graph Editing Distance}
\acrodef{llm}[LLM]{Large Language Model}
\acrodef{rlhf}[RLHF]{Reinforcement Learning from Human Feedback}
\acrodef{iclhf}[ICLHF]{In-Context Learning from Human Feedback}

\title{\LARGE \bf
In-situ Value-aligned Human-Robot Interactions \\
with Physical Constraints
}

\author{Hongtao Li$^{1,2}$, Ziyuan Jiao$^{2}$, Xiaofeng Liu$^{1,\dagger}$~\IEEEmembership{Senior Member,~IEEE}, Hangxin Liu$^{2,\dagger}$, Zilong Zheng$^{2,\dagger}$
\thanks{$^1$ College of Artificial Intelligence and Automation, Hohai University, Changzhou, China}%
\thanks{$^2$ State Key Laboratory of General Artificial Intelligence, BIGAI, Beijing, China}%
\thanks{This work is done during H. Li's internship at BIGAI. $^\dagger$ indicates the corresponding authors. E-mails: \texttt{\{lihongtao, jiaoziyuan, liuhx, zlzheng\}@bigai.ai, xfliu@hhu.edu.cn}}%
}

\begin{document}

\maketitle
\thispagestyle{empty}
\pagestyle{empty}

\begin{abstract}

Equipped with \acp{llm}, human-centered robots are now capable of performing a wide range of tasks that were previously deemed challenging or unattainable. However, merely completing tasks is insufficient for cognitive robots, who should learn and apply human preferences to future scenarios. In this work, we propose a framework that combines human preferences with physical constraints, requiring robots to complete tasks while considering both. Firstly, we developed a benchmark of everyday household activities, which are often evaluated based on specific preferences. We then introduced \ac{iclhf}, where human feedback comes from direct instructions and adjustments made intentionally or unintentionally in daily life. Extensive sets of experiments, testing the \ac{iclhf} to generate task plans and balance physical constraints with preferences, have demonstrated the efficiency of our approach.

\end{abstract}

\section{Introduction}

\begin{figure}[t!]
    \centering
    \includegraphics[width=\linewidth]{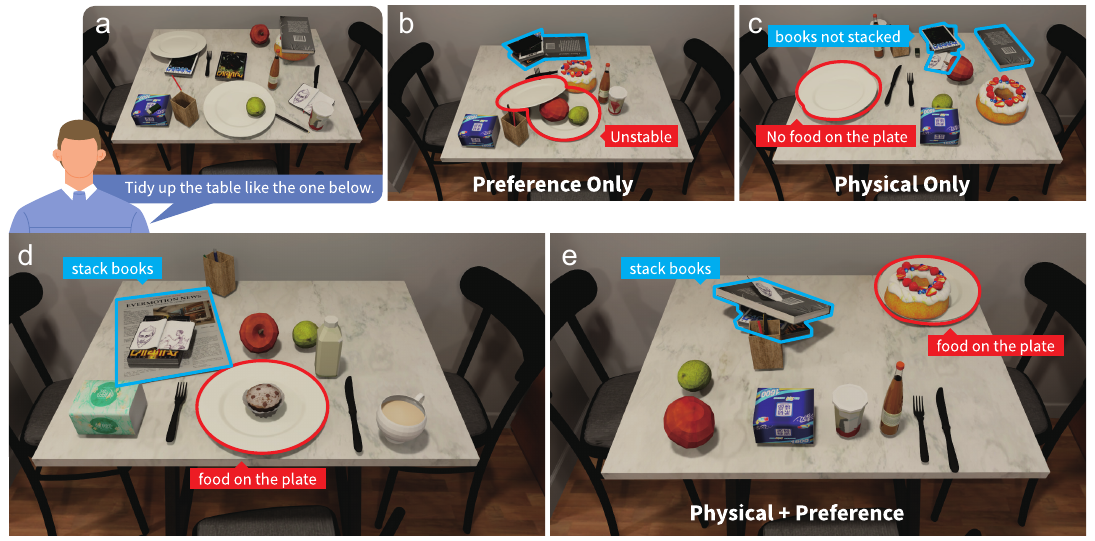}
    \caption{\textbf{An example task of robots tidying up a table with human preferences.} (a) The messy table that needs to be cleared. (b) Considering human preferences without regard to physical constraints would result in unrealistic behavior. (c) Considering physical constraints alone would fail to meet human expectations. (d) Example of human preferences. (e) Only by balancing both, that is, behaving following human preferences while adhering to physical constraints, can the task be satisfactorily completed.}
    \label{fig:motivation}
    \vspace{\myvspace}
\end{figure}

Equipping robots, especially service robots, with the ability to consider personalized human preferences is a challenging task. On the one hand, this is due to the subjectivity and diversity of human preferences, and on the other hand, the physical constraints of the objective world limit the realization of preferences. Imagine a scenario where robots tidy up a table, as shown in \cref{fig:motivation}, where humans expect the robot to tidy up according to their preferences. Therefore, it is inappropriate to consider preferences without regard to physical constraints, or physical constraints only, but only to take both into account, i.e., behaving in accordance with human preferences while adhering to physical constraints.

One of the challenges is to learn human preferences, with existing methods mainly including learning by pairwise comparison~\cite{furnkranz2003pairwise, furnkranz2010preference} and learning by \acp{llm}~\cite{wu2023tidybot,zha2024distilling,ma2024eureka}. The former typically simplifies complex preferences into a ranking function, which assigns a partial order to multiple outputs of the model. Such methods are easy to implement and have numerous applications in recommendation systems~\cite{zhao2016user,he2017category}, human-robot interaction~\cite{cakmak2011human}, natural language processing~\cite{christiano2017deep,hejna2023few}, and more. However, they require collecting a large amount of annotated data from humans~\cite{ji2023beavertails,kirstain2023pick,wu2023human}, so what is learned reflects the shared preferences rather than individualized preferences~\cite{bakker2022fine}. Additionally, this form of comparison itself also sacrifices the diversity of human preferences. The latter arises from the powerful text processing and common-sense reasoning capabilities exhibited by \acp{llm} trained on massive datasets~\cite{wei2022chain,yao2023tree}. For example, TidyBot~\cite{wu2023tidybot} utilizes the summarization capabilities of \acp{llm} to infer human preferences for tidying up a room, while DROC~\cite{zha2024distilling} learns human preferences from online human language corrections.

Another major challenge is the integration of learned human preferences and physical constraints. Physical constraints manifest in both task planning~\cite{dantam2018incremental} and motion planning~\cite{marcucci2023motion} of robots. In this paper, we only consider the physical constraints of the task planning, and the human preferences associated with it only involve the outcomes of planning. Although it is a subset of the original problem, it remains a challenging task. Due to the inherent, universal, and omnipresent nature of physical constraints, they are typically predefined in the form of hand-written rules, such as the domain description in \ac{pddl}~\cite{mcdermott20001998,fox2003pddl2}, or environment modeling in reinforcement learning~\cite{reda2020learning}. On the other hand, personalized human preferences are often unpredictable, ambiguous, and diverse. The heterogeneity between these two demands an appropriate way of combining them~\cite{jorge2008planning}.

Traditional task planning methods, such as those based on \ac{pddl} or scene graphs, require converting preferences into a form recognizable by planners, such as preference predicates~\cite{canal2019adapting} or constraints~\cite{kim2017collaborative}. However, this conversion often sacrifices the diversity of preferences. Another approach is to use \acp{llm} to convert text-based human preferences into a reward function in reinforcement learning~\cite{yu2023language,ma2024eureka} and learn physical constraints through a large number of trial-and-error. However, the intrinsic nature of reinforcement learning requires human preferences, serving as reward signals, to be strongly goal-oriented for specific tasks~\cite{eschmann2021reward}, thus making it less suited for learning diverse preferences.

To address the aforementioned issues, we first proposed a set of household benchmarks, collecting tasks with strong personal preferences. Then, we introduced the \acf{iclhf} algorithm, which aims to combine \acp{llm}' preference learning capability with the ability to learn from feedback. In this approach, the \ac{llm} functions like a reinforcement learning policy model, learning human preferences through in-context learning~\cite{dong2022survey,min2022rethinking}. Feedback is provided in textual form, combining physical constraints and human preferences. Finally, the structure like \ac{rlhf} of the algorithm allows for balancing physical feedback and preference feedback to generate suitable solutions. Meanwhile, the employment of in-context learning avoids training an excess of parameters for \acp{llm}, maximizing its inferential prowess as much as possible, thus enabling in-situ personalized preference learning. To achieve generalization, learned human preferences can be easily combined into a hierarchical structure, with higher-level preferences being more adaptive. Considering the powerful capability of traditional algorithms in handling intricate manipulation tasks, we integrated a customized version of POG\cite{jiao2022sequential}, an algorithm for efficient sequential manipulation planning on scene graphs, to enhance the preliminary plans generated by the \ac{llm} planner. Consequently, the final task plan incorporates more comprehensive geometric spatial information, thereby ensuring seamless transitions to the motion planner.

We conducted numerous experiments to validate the effectiveness of the \ac{iclhf} algorithm in learning human preferences and combining them with physical constraints. Finally, real robot experiments demonstrate the validation of the approach on robotic hardware. The contribution of our work can be summarized as follows:
\begin{itemize}
    \item We present a benchmark on household activities whose evaluation is based on personalized preferences.
    \item We introduce the \ac{iclhf} algorithm that learns human preferences in situ and combines them with physical constraints to accomplish the task.
    \item We conduct large-scale experiments to validate the effectiveness of \ac{iclhf}, and real-world robot experiments to demonstrate its practicality.
\end{itemize}

\subsection{Related Works}
\subsubsection{LLMs for Robots}
As \acp{llm} trained on massive amounts of data exhibit powerful common-sense reasoning capabilities~\cite{wei2022chain,yao2023tree}, a significant amount of work focuses on utilizing them for robotic task planning~\cite{yu2023language,rana2023sayplan,singh2023progprompt}. These works can be categorized based on the format of \acp{llm}' output into two types: action primitives based~\cite{rana2023sayplan,wu2023tidybot} methods and code-based~\cite{yu2023language,singh2023progprompt,ma2024eureka} methods. Methods based on action primitives guide \acp{llm} to generate corresponding sequences of action primitives based on different task prompts, while methods based on code aim to leverage the programming abilities of \acp{llm} by providing specific API instructions to output execution plans~\cite{singh2023progprompt} or objective functions~\cite{yu2023language,ma2024eureka} represented in code format.

\subsubsection{Task Planning}
Traditional task planning in robotics mainly includes planning with symbols~\cite{canal2019adapting,kim2017collaborative} and planning with scene graphs~\cite{jiao2022sequential,rana2023sayplan,zhu2021hierarchical}. The utilization of symbols for robotic task planning derives from earlier planning problems, where algorithms such as \ac{pddl}~\cite{mcdermott20001998,fox2003pddl2} standardize artificial intelligence planning. 3D scene graph~\cite{armeni20193d,chang2021comprehensive} emerges as a formidable tool for scene modeling and makes many graph operations possible due to the graph structure, such as graph edit distance~\cite{jiao2022sequential} and graph neural networks~\cite{yang2018graph}. However, it also has some issues whereby even small environments can contain hundreds of objects and complex relationships between them~\cite{agia2022taskography}.

\subsection{Overview}
We organize the remainder of this paper as follows. \cref{sec:method} describes the underlying framework and the \ac{iclhf} algorithm constructed on it. \cref{sec:experiments} presents the benchmark used in this paper, and exhaustive experiments are conducted on both simulation and real environments to validate the effectiveness of the \ac{iclhf} algorithm. Finally, we conclude the paper in \cref{sec:conclusion}.

\begin{figure*}[th!]
    \centering
    \includegraphics[width=\linewidth]{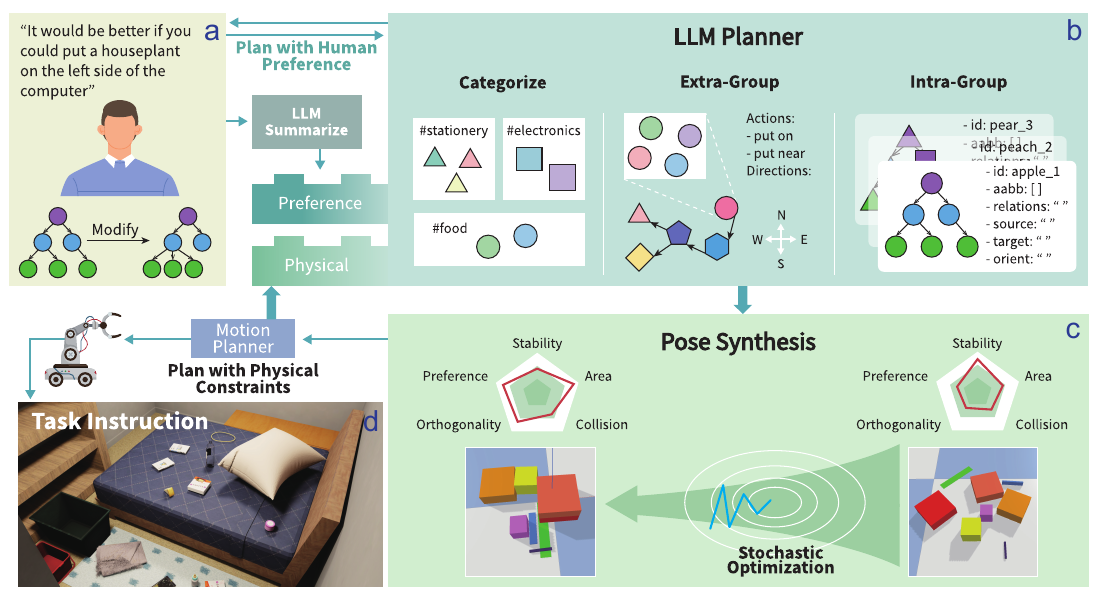}
    \caption{\textbf{Dual loop of ICLHF}. (a-b) In the upper loop, humans give instructions or adjust the plan according to their preferences, which are then analyzed and integrated by the \ac{llm} into human preferences. (b) In the lower loop, \ac{llm} receives the initial state and task instruction as input and outputs the object attributes and relationships between objects represented by the scene graph. (c) Then, the pose synthesizer processes this scene graph and generates specific positions and rotations for the object. (d) Finally, the motion planner generates the robot's motion trajectory based on this information. After executing in the environment, the robot receives physical feedback, which is then injected into the \ac{llm} planner for subsequent planning. The whole process continues until the generated plan conforms to the physical constraints and human preferences.}
    \label{fig:detail}
    \vspace{\myvspace}
\end{figure*}

\section{Method} \label{sec:method}
\subsection{Framework}
The problem we consider is similar to the traditional \ac{mdp}, where the state space $\mathcal{S}$ encompasses all possible states of objects, including poses, intrinsic properties, and more. The action space $\mathcal{A}$ consists of predefined action primitives, such as group, put on, and slice. Since it is often challenging~\cite{ma2024eureka,yu2023language} to accurately describe physical constraints and human preferences using the traditional scalar reward function $R: \mathcal{S} \times \mathcal{A} \to \mathbb{R}$, we employ a novel text-based form of reward to integrate both, defined as $R^*: \mathcal{S} \times \mathcal{A} \to \mathbb{F}^\text{text}$. In the simulation, physical constraints are typically provided by simulators, whereas in the real world, constraints and preferences are provided by human observers either in the form of speech (which can be converted to text) or directly as text.

Specifically, environmental feedback consists of two parts. First is the execution of actions, for example, any placement actions within the container before it is opened are considered failures. Second is the consequences of actions, such as collisions or collapses caused by the operation. Based on such feedback, an intelligent agent can generate plans with higher physical feasibility. Human feedback also includes two parts. First is direct preference instructions from humans regarding unsatisfactory aspects of the execution process, demanding the agent to respond promptly and generate plans that align with human preferences. Second is the adjustments made by humans in daily life based on their preferences, which are more implicit compared to the first type of feedback, sometimes even stemming from subconscious human behavior. Inductive learning of these preferences poses a challenging task. Traditional algorithms~\cite{canal2019adapting,kim2017collaborative} lack effective learning methods for this, but \acp{llm} can summarize corresponding human preferences and apply them to subsequent tasks~\cite{wu2023tidybot}.

\subsection{ICLHF}
We propose the \ac{iclhf} algorithm, which is capable of learning human preferences in situ and combining them with physical constraints to accomplish tasks. It consists of two parts: the \ac{llm} planner and the object pose synthesizer. Taking task instructions and initial states as input, the \ac{llm} planner performs in-context learning based on corresponding prompts and recorded human preferences, and outputs execution sequences along with the goal scene graph. The output scene graph is a rough version, representing objects and their attributes with nodes and carrying the relationships between objects with edges. The pose synthesizer then takes this goal scene graph as input and further synthesizes more specific object information, such as position and rotation, based on object attributes and relationships between objects. Additionally, the pose synthesizer can conduct a preliminary physical feasibility analysis on scene graphs, such as object collisions, and provide feedback on the physical aspects.

Throughout the entire process, humans can provide modification suggestions or make adjustments based on their preferences at any time. The algorithm can capture these human preferences and apply them to subsequent planning. Additionally, to utilize human preferences more efficiently, the algorithm performs periodic introspection to extract higher-level human characteristics from lower-level human preferences, typically when reaching the maximum context length of \acp{llm}. The overall process is illustrated in \cref{fig:detail}.

\subsubsection{LLM as Task Planner}
The task planner adopts a top-down processing logic. Given the textual description of objects, first classify the objects. Assuming this step generates $N$ categories, a total of $N+1$ directed acyclic graphs will be obtained. Objects of the same category are considered as nodes within the same graph, while these $N$ categories themselves form a graph with $N$ nodes.

Next, consider the placement between groups, mainly involving actions of \textit{put on} and \textit{put near}, as well as optional orientation indications, which will provide a global placement for the $N$ categories divided in the previous step.

Finally, groups with more than two objects will undergo more detailed operations, with the types varying according to tasks. Taking tidying up a table as an example, operations will include \textit{put on}, \textit{put in}, \textit{open}, \textit{close}, and so on. These operations either alter the relationship between objects, such as \textit{put on} and \textit{put in}, or change the state of a single object, like \textit{open} and \textit{close}.

In each of the above processes, physical feedback and preference feedback can be promptly injected to influence subsequent planning. With physical feedback, the agent can modify parts of the plan that are not executable or unrealistic, while preference feedback will affect future planning. During the modification process, potential human preferences also need to be considered. Therefore, unlike the error handling mechanism proposed in DROC~\cite{zha2024distilling}, which restricts retrievable history to four categories, we track the source of relationships in the scene graph that lead to errors and use them together with the relationships of neighbors as contextual input to regenerate an overall plan that better aligns with human preferences. Additionally, when the stored preferences reach the maximum token length allowed by \acp{llm}, the planner will conduct a profile, aiming to extract more generalized features from trivial preferences.

\subsubsection{POG as Pose Synthesizer}
In contrast to the top-down logic of the task planner, the pose synthesizer adopts a bottom-up approach. It first analyzes the placement of objects within each group, then treats them as a whole to generate a total of $N$ placement configurations for all groups based on their orientation and relationships with other groups.

Specifically, for the symbolic relationships generated by the task planner, we use stochastic optimization in POG~\cite{jiao2022sequential} to determine the geometric information of the objects. To reduce computational complexity, the oriented bounding box is used instead of objects during calculations, and then the results are mapped back to the respective objects. In addition to the original objective function used in POG~\cite{jiao2022sequential}, the following additional objectives have been added.

\begin{equation} \label{eq:1}
    \mathcal{L}_\text{manhattan} := \sum_{l \in \mathcal{G}} \mathbf{1}_{|l| > 1} \sum_{\mathbf{m}, \mathbf{n} \in l} \|\mathbf{m} - \mathbf{n}\|_1
\end{equation}
\begin{equation} \label{eq:2}
    \mathcal{L}_\text{area} := \mathcal{L}_\text{manhattan} + \sum_{l \in \mathcal{G}} \mathbf{1}_{|l| > 1} R(\mathbf{x}^l) \cdot R(\mathbf{y}^l)
\end{equation}
\begin{equation} \label{eq:3}
    \mathcal{L}_\text{orth} := \sigma^2 (\boldsymbol{\theta})
\end{equation}
where $l$ denotes the depth of nodes in the scene graph and $\mathbf{1}$ denotes the indicator function. The $\mathbf{m}, \mathbf{n}$ in \cref{eq:1} denote the 3D coordinates of nodes. In \cref{eq:2}, $R$ denotes the range, i.e. $R(\mathbf{x}) = \mathbf{x}_\text{max} - \mathbf{x}_\text{min}$, and $x, y$ denote the x-axis and y-axis coordinates of the node, respectively. In \cref{eq:3}, $\boldsymbol{\theta}$ is the intersection angle between the main axis of symmetry and the x-axis for each object.

\cref{eq:2} is primarily aimed at reducing the distance between objects and consists of two parts. Firstly, it constrains the distance between every two objects using Manhattan distance, as \cref{eq:1} shows, which, combined with \cref{eq:3}, can make the arrangement of objects neat and in line with human preferences. Secondly, it constrains multiple objects to make them more compact as a whole. \cref{eq:3} aims to reduce deviations between the main axes of symmetry of objects. These metrics reflect more fundamental and general preferences, which, when combined with individual unique preferences, can model human preferences from multiple dimensions. Additionally, we extract objectives from POG~\cite{jiao2022sequential} regarding stability and collisions to form quantitative metrics. It is worth noting that the stability and collision functions can also provide preliminary physical feedback.

\section{Experiments} \label{sec:experiments}
Our experiment needs to answer the following questions:
\begin{enumerate}
    \item Why choose in-context learning to learn human preferences, and what advantages does it have over directly using \acp{llm} for learning?
    \item Can the \ac{llm} planner generate more detailed information based on the symbolic relationships between objects, such as object positions and rotations?
    \item Can the \ac{iclhf} plan in a way that aligns with preferences while also adhering to physical constraints, and can it generalize to new scenarios with minimal effort?
    \item How practical is the \ac{iclhf} algorithm, and can it be applied to real robots?
\end{enumerate}

We conducted numerous experiments to answer the aforementioned questions. The remainder of this section is organized as follows. First, in \cref{sec:benchmark}, we introduce a benchmark comprising common household tasks, each requiring specific preferences as evaluation criteria. Then, in \cref{sec:exp_1}, experiments are conducted to validate the ability of in-context learning as a preference learner. Following that, \cref{sec:exp_2} compares the ability of \acp{llm} and the traditional algorithm, namely POG~\cite{jiao2022sequential}, in generating object geometric information. \cref{sec:exp_3} carries out extensive experiments to analyze the ability of \ac{iclhf} to balance physical constraints and human preferences, as well as its generalization. Finally, \cref{sec:exp_4} validates the effectiveness of \ac{iclhf} in real robot environments. Throughout the experiments, we utilized GPT-3.5 Turbo as the \ac{llm} planner.

\subsection{Benchmark} \label{sec:benchmark}
\begin{figure}[ht]
    \centering
    \includegraphics[width=\linewidth]{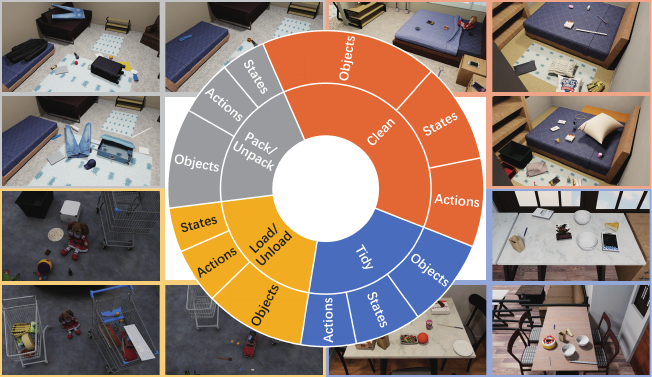}
    \caption{Our benchmark covers four common types of tasks in household chores, the execution of which typically involves distinct human preferences, namely tidying up, cleaning, packing, and loading.}
    \label{fig:benchmark}
    \vspace{\myvspace}
\end{figure}

From Behavior-1K~\cite{li2023behavior}, a collection of household activities matching human needs based on a large number of surveys, we filtered out four categories of tasks, the execution of which typically involves distinct human preferences, namely tidying up, cleaning, packing/unpacking, and loading/unloading, as shown in \cref{fig:benchmark}.

There are a total of 15 activities and 22 different scenarios in the benchmark, involving a total of 1596 objects. \cref{tab:activity_attributes} shows the attributes of tasks in different types, each obtained by averaging and rounding across all available scenarios.

\begin{table}[h]
    \centering
    \caption{The four types of activities and their attributes}
    \small
    \begin{tabular}{c|c|c|c|c}
    \toprule[1pt]
    \textbf{Activity Type} &\textbf{Objects} &\textbf{States} &\textbf{Actions} &\textbf{Amount} \\
    \hline
    tidy &14 &10 &10 &150 \\
    clean &21 &16 &12 &100 \\
    pack/unpack &19 &11 &11 &80 \\
    load/unload &17 &12 &11 &80 \\
    \bottomrule[1pt]
    \end{tabular}
    \label{tab:activity_attributes}
\end{table}

For each type of task, we provide default human preferences. For instance, the human preference is ``I prefer everything to be laid flat on the table rather than stacked together'' for the task of tidying up tables. Each preference is carefully selected to ensure that there is at least one solution that aligns with the preference in the current context, while also being as general as possible to influence other types of tasks in specific scenarios.

Additionally, the benchmark considers the impact of preferences on the physical difficulty of task completion, implicitly increasing or decreasing constraints by adjusting the physical contact between objects. For example, the default preference for tidying up tables avoids stacking objects, making it easier to execute, while the default preference for unloading cars suggests placing objects in the same container, greatly reducing the feasible domain of the task. The evaluation of human preferences consists of two parts: subjective scoring and objective scoring. The subjective scoring is performed by selected participants, who rate the final RGB image from 0 to 10 based on given preferences. The objective scoring is calculated by selecting different features with varying weights according to specific preferences.

\subsection{Symbolic Spatial Relationship Experiments} \label{sec:exp_1}
\subsubsection{Settings}
To test the ability of in-context learning to learn human preferences, we conducted experiments using tidying up tables as an example on the PyBullet~\cite{coumans2021} platform, involving 5 to 10 objects. This test includes two aspects. First, the ability of the algorithm to extract human preferences from modifications in object relationships, and second, the understanding and application of human preferences. To standardize the output format, in methods that do not involve in-context learning, only content related to preference learning has been removed.

Considering the inherent ability of \acp{llm} to process semantic information, we categorize object types used in experiments into those containing semantic information, namely everyday items, and those lacking semantic information, which mainly include boxes and cylinders.

The evaluation criteria are divided into three levels: scene graph, action sequence, and preference. The scene graph includes stability and area, the action sequence includes execution efficiency and feasibility, and the preference includes learning and application. In the evaluation of the scene graph, the stability cost function is defined as
\begin{equation} \label{eq:4}
    \mathcal{L}_\text{stab} := \frac{\sum_o \text{Mass}_o \cdot \Vert \text{CoM}_o \Vert_2 + \Vert \sum_o \text{Mass}_o \cdot \text{CoM}_o \Vert_2}{\sum_o \text{Mass}_o}
\end{equation}
where $\text{Mass}_o$ denotes the mass of object $o$, and $\text{CoM}_o$ represents the center of mass of object $o$ relative to the base object. The area cost function is defined as \cref{eq:1,eq:2}. The corresponding scores are scaled and transformed into a range of 0 to 10 through min-max normalization. The execution efficiency of the action sequence is inversely proportional to the length of the task plan, and feasibility includes logical feasibility and physical feasibility. Logical feasibility analyzes whether the inherent logic of the plan is correct, including the format of instructions, while physical feasibility analyzes whether the plan can be successfully executed physically. Preference learning utilizes SentenceTransformers~\cite{reimers-2019-sentence-bert} to measure the cosine similarity between learned preferences and the ground truth. The application of preferences tests the algorithm's understanding of preferences by evaluating its application to new scenarios. The overall score is calculated as the average of the subjective and objective scores from the benchmark. The scaling method for the objective score follows the same approach as the stability.

\subsubsection{Results}
\begin{table}[h]
    \centering
    \caption{Results of task planning without semantic information (average across 5 to 10 objects)}
    \small
    \begin{tabular}{cccc}
    \toprule[1pt]
    \multicolumn{2}{c}{\textbf{Criteria}} & \textbf{With ICL} & \textbf{Without ICL} \\
    \hline
    \multirow{2}{*}{Goal} & Stability $\uparrow$ &7.18 &\textbf{7.28} \\ & Area $\downarrow$ &\textbf{8.45} &8.61 \\
    \hline
    \multirow{2}{*}{Sequence} & Length $\downarrow$ &14.81 &\textbf{14.67} \\ & Feasibility $\uparrow$ &\textbf{12.98} &8.44 \\
    \hline 
    \multirow{2}{*}{Preference} & Learn $\uparrow$ &\textbf{0.95} &0.47 \\ & Apply $\uparrow$ &\textbf{85.48} &66.67 \\
    \bottomrule[1pt]
    \end{tabular}
    \label{tab:no_semantic_llm_task_plan}
\end{table}

\begin{figure}[h]
    \centering
    \includegraphics[width=\linewidth]{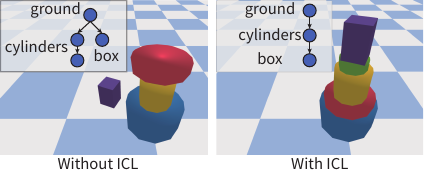}
    \caption{The example results generated with or without in-context learning, where humans prefer mixing all objects together.}
    \label{fig:sym_pybullet}
    \vspace{\myvspace}
\end{figure}

\begin{figure*}[!th]
    \centering
    \includegraphics[width=\linewidth]{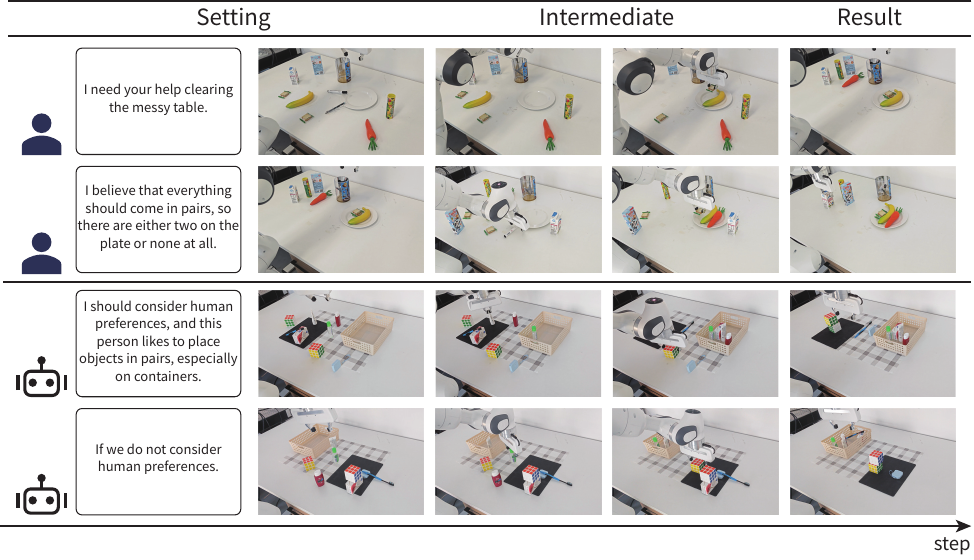}
    \caption{The results of two sets of real robots tidying up a table. The first was configured with zero-shot learning, while the second used one-shot learning. Results without preference were also provided for comparison.}
    \label{fig:experiment_real}
    \vspace{\myvspace}
\end{figure*}

\begin{figure}[!ht]
    \centering
    \includegraphics[width=\linewidth]{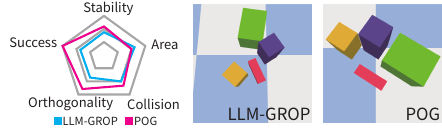}
    \caption{The example of using LLM-GROP and POG to generate object poses.}
    \label{fig:geo_pybullet}
    \vspace{\myvspace}
\end{figure}

\cref{tab:no_semantic_llm_task_plan} displays the corresponding results when there is no semantic information. In this experiment, 5 to 10 objects are randomly sampled within appropriate ranges of categories and sizes. When the number of objects in a category is less than one-third of the total, or greater than two-thirds, the preference is to mix boxes and cylinders, meaning there exist instances of one category of objects placed on top of another. In all other cases, the preference is to separate boxes and cylinders. From \cref{tab:no_semantic_llm_task_plan}, it can be seen that using in-context learning greatly improves the feasibility of generating plans, as well as preference learning and application. \cref{fig:sym_pybullet} shows visual examples of scenarios requiring mixed objects.

\cref{tab:semantic_llm_task_plan} presents the results with semantic information. When identical objects are present, the preference is set to disallow stacking identical objects together. From \cref{tab:semantic_llm_task_plan}, it can be observed that the overall feasibility of the plan is improved when using everyday objects, while the use of in-context learning still enhances preference learning and application capabilities. Additionally, we observed that as the number of objects increases, the logical feasibility of the plans generated by the \ac{llm} significantly decreases, with frequent occurrences of hallucinations and inconsistencies. This issue is particularly evident in methods that do not involve in-context learning, where the \ac{llm} often generates nonexistent actions or operation logic in an attempt to handle human preferences.

\begin{table}[h]
    \centering
    \caption{Results of task planning with semantic information (average across 5 to 10 objects)}
    \small
    \begin{tabular}{cccc}
    \toprule[1pt]
    \multicolumn{2}{c}{\textbf{Criteria}} & \textbf{With ICL} & \textbf{Without ICL} \\
    \hline
    \multirow{2}{*}{Goal} & Stability $\uparrow$ &\textbf{8.69} &8.62 \\ & Area $\downarrow$ &8.02 &\textbf{7.98} \\
    \hline
    \multirow{2}{*}{Sequence} & Length $\downarrow$ &\textbf{12.32} &12.65 \\ & Feasibility $\uparrow$ &\textbf{26.45} &10.74 \\
    \hline 
    \multirow{2}{*}{Preference} & Learn $\uparrow$ &\textbf{0.86} &0.39 \\ & Apply $\uparrow$ &\textbf{93.33} &89.02 \\
    \bottomrule[1pt]
    \end{tabular}
    \label{tab:semantic_llm_task_plan}
\end{table}

\subsection{Geometric Spatial Relationship Experiments} \label{sec:exp_2}

\subsubsection{Settings}
We aim to have the \ac{llm} generate more precise geometric spatial relations based on symbolic spatial relationships using the algorithm proposed in LLM-GROP~\cite{ding2023task}, and we compare these results with traditional optimization-based algorithms. The experimental setup, based on LLM-GROP~\cite{ding2023task}, involves a service robot tasked with arranging a dining table. For testing, we sample 3 to 5 objects from 7 categories, totaling 26.

The evaluation criteria for this experiment include success rate and orthogonality. The success rate is defined as whether the method can place all objects on the table without collisions. Orthogonality is defined as in \cref{eq:3}.

\subsubsection{Results}
\cref{tab:geometric_plan} shows the results. Orthogonality is transformed into scores using min-max scaling. It can be observed that even with only 3 to 5 objects, the success rate of LLM-GROP~\cite{ding2023task} is below 60\%, while POG~\cite{jiao2022sequential} achieves 100\%. Additionally, POG~\cite{jiao2022sequential} outperforms in orthogonality, indicating a neater placement of objects. \cref{fig:geo_pybullet} illustrates examples of object poses generated using LLM-GROP~\cite{ding2023task} and POG~\cite{jiao2022sequential}, respectively.

\begin{table}[h]
    \centering
    \caption{Results of geometric spatial planning (average across 3 to 5 objects)}
    \small
    \begin{tabular}{c|c|c}
    \toprule[1pt]
    \textbf{Criteria} &\textbf{LLM-GROP} &\textbf{POG} \\
    \hline
    Orthogonality Score ($\cdot/10$) &5.45 &\textbf{7.04} \\
    Success Rate (\%) &56.67 &\textbf{100} \\
    \bottomrule[1pt]
    \end{tabular}
    \label{tab:geometric_plan}
\end{table}

\subsection{Simulation Experiments} \label{sec:exp_3}

\subsubsection{Settings}
We conducted ablation experiments on the OmniGibson~\cite{li2023behavior} platform to validate the \ac{iclhf} algorithm's ability to plan in accordance with human preferences while adhering to physical constraints. The experiments are based on the benchmark we previously proposed and are divided into four categories: tidying up, unloading, unpacking, and cleaning, each with corresponding human preferences. To enhance the complexity of the experiment, we integrated various human preferences from previous tasks into a room tidying task. This enabled us to analyze how the algorithm balances complex and diverse preferences in more realistic scenarios. Additionally, we observed that \ac{llm} itself possesses many common human preferences, so the preferences used in the experiments have distinct personalities. The number of objects in the experiments ranges from 5 to 15, with their categories and poses sampled within appropriate ranges. The method for evaluating preferences is similar to the previous one, supplemented with quantitative scores.

\subsubsection{Results}
\begin{figure}[!t]
    \centering
    \includegraphics[width=\linewidth]{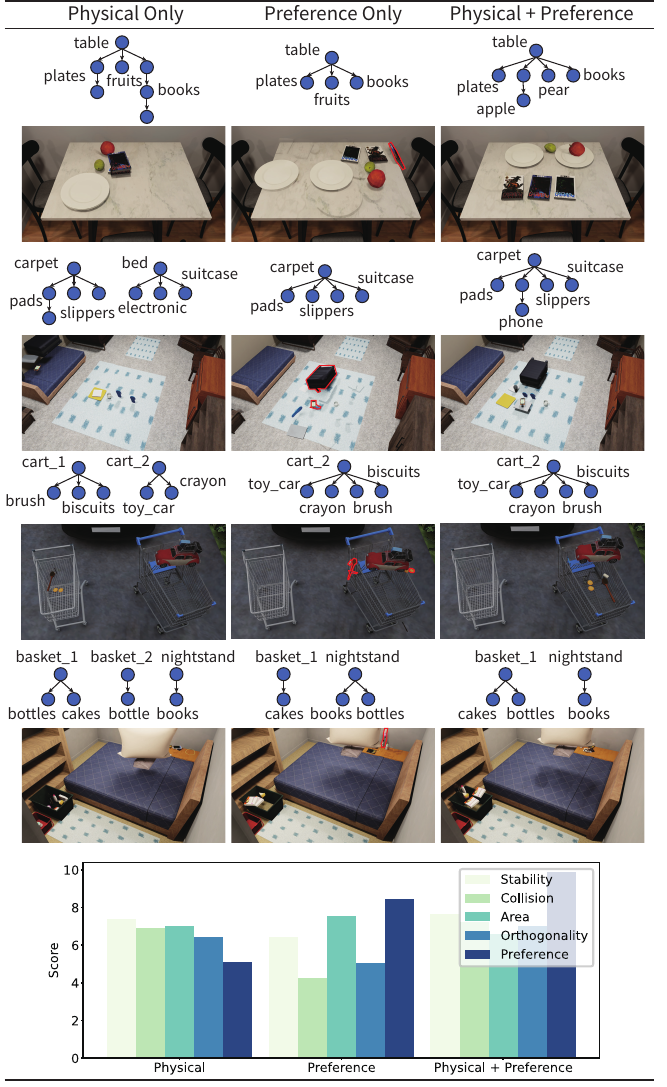}
    \caption{The results of the ablation experiments include visualizations of selected scenes, supplemented by quantitative data analysis. Preferences for tidying involve laying objects flat on the table. Preferences for unloading entail placing all items in the same cart, and unpacking preferences dictate avoiding placing items unrelated to sleeping on the bed. The \ac{iclhf} algorithm, which integrates both physical constraints and human preferences, can generate plans that are physically feasible while also aligning with human preferences.}
    \label{fig:simulation}
    \vspace{\myvspace}
\end{figure}

\cref{fig:simulation} presents the results of the experiments, visualizing some scenarios and supplemented with quantitative scores. The fourth row of RGB images depicts a scene that combines the preferences from the scenes in the first three rows. It can be seen that plans generated solely under physical constraints do not meet specific human preferences, while considering preferences alone may result in impractical plans. Only the \ac{iclhf} algorithm, which simultaneously considers both physical constraints and human preferences, is capable of addressing this challenge. Additionally, the algorithm demonstrates strong generalization abilities, meaning previously learned human preferences can be applied to unknown scenarios.

The computational complexity primarily involves \ac{llm} and the pose synthesis module. The latter's efficiency is improved through the use of oriented bounding boxes and parallel processing of different groups, averaging 0.6 seconds in current experiments.

\subsection{Real Robot Experiments} \label{sec:exp_4}
Lastly, we conducted manipulation experiments using a Franka Research 3 manipulator, where the task was to tidy up a table. Initially, the robotic arm followed physical constraints to tidy up the table efficiently. Subsequently, human preferences were explicitly expressed, and the robotic arm adjusted its actions accordingly. When faced with a new scenario, the robotic arm planned actions based on previously learned human preferences while strictly adhering to physical constraints. As a comparison, results without considering human preferences were provided for evaluation. The experimental results are shown in \cref{fig:experiment_real}, demonstrating the robot's excellent task completion and adaptability.

\section{Conclusion} \label{sec:conclusion}
In this paper, we introduce a dual-loop planning framework that integrates physical constraints and human preferences, offering a novel human-in-the-loop paradigm. Based on this framework, we propose the \acf{iclhf} algorithm, which can learn human preferences in situ and make plans that adhere to physical constraints while aligning with preferences. To validate the effectiveness of the proposed algorithm, we introduce a novel benchmark that incorporates personalized preferences into the evaluation process. We conduct extensive experiments to verify the capabilities of the \ac{iclhf} algorithm across various aspects. Finally, real robot experiments demonstrate its practicality in robotic hardware.





\section*{Acknowledgment}

The authors thank the reviewers for their insightful suggestions to improve the manuscript. This work presented herein is supported by the National Natural Science Foundation of China (62376031).


\bibliographystyle{IEEEtran}
\bibliography{references}

\begin{thebibliography}{10}
\providecommand{\url}[1]{#1}
\csname url@rmstyle\endcsname
\providecommand{\newblock}{\relax}
\providecommand{\bibinfo}[2]{#2}
\providecommand\BIBentrySTDinterwordspacing{\spaceskip=0pt\relax}
\providecommand\BIBentryALTinterwordstretchfactor{4}
\providecommand\BIBentryALTinterwordspacing{\spaceskip=\fontdimen2\font plus
\BIBentryALTinterwordstretchfactor\fontdimen3\font minus
  \fontdimen4\font\relax}
\providecommand\BIBforeignlanguage[2]{{%
\expandafter\ifx\csname l@#1\endcsname\relax
\typeout{** WARNING: IEEEtran.bst: No hyphenation pattern has been}%
\typeout{** loaded for the language `#1'. Using the pattern for}%
\typeout{** the default language instead.}%
\else
\language=\csname l@#1\endcsname
\fi
#2}}

\bibitem{furnkranz2003pairwise}
J.~F{\"u}rnkranz and E.~H{\"u}llermeier, ``Pairwise preference learning and
  ranking,'' in \emph{Machine Learning: ECML 2003}.\hskip 1em plus 0.5em minus
  0.4em\relax Berlin, Heidelberg: Springer Berlin Heidelberg, 2003, pp.
  145--156.

\bibitem{furnkranz2010preference}
------, ``Preference learning and ranking by pairwise comparison,'' in
  \emph{Preference Learning}.\hskip 1em plus 0.5em minus 0.4em\relax Berlin,
  Heidelberg: Springer Berlin Heidelberg, 2011, pp. 65--82.

\bibitem{wu2023tidybot}
J.~Wu, R.~Antonova, A.~Kan, M.~Lepert, A.~Zeng, S.~Song, J.~Bohg,
  S.~Rusinkiewicz, and T.~Funkhouser, ``Tidybot: Personalized robot assistance
  with large language models,'' \emph{Autonomous Robots}, vol.~47, no.~8, pp.
  1087--1102, 2023.

\bibitem{zha2024distilling}
L.~Zha, Y.~Cui, L.-H. Lin, M.~Kwon, M.~G. Arenas, A.~Zeng, F.~Xia, and
  D.~Sadigh, ``Distilling and retrieving generalizable knowledge for robot
  manipulation via language corrections,'' in \emph{International Conference on
  Robotics and Automation (ICRA)}, 2024, pp. 15\,172--15\,179.

\bibitem{ma2024eureka}
Y.~J. Ma, W.~Liang, G.~Wang, D.-A. Huang, O.~Bastani, D.~Jayaraman, Y.~Zhu,
  L.~Fan, and A.~Anandkumar, ``Eureka: Human-level reward design via coding
  large language models,'' in \emph{International Conference on Learning
  Representations (ICLR)}, 2024.

\bibitem{zhao2016user}
Z.~Zhao, H.~Lu, D.~Cai, X.~He, and Y.~Zhuang, ``User preference learning for
  online social recommendation,'' \emph{IEEE Transactions on Knowledge and Data
  Engineering}, vol.~28, no.~9, pp. 2522--2534, 2016.

\bibitem{he2017category}
J.~He, X.~Li, and L.~Liao, ``Category-aware next point-of-interest
  recommendation via listwise bayesian personalized ranking,'' in
  \emph{International Joint Conference on Artificial Intelligence (IJCAI)},
  2017, pp. 1837--1843.

\bibitem{cakmak2011human}
M.~Cakmak, S.~S. Srinivasa, M.~K. Lee, J.~Forlizzi, and S.~Kiesler, ``Human
  preferences for robot-human hand-over configurations,'' in
  \emph{International Conference on Intelligent Robots and Systems (IROS)},
  2011, pp. 1986--1993.

\bibitem{christiano2017deep}
P.~F. Christiano, J.~Leike, T.~Brown, M.~Martic, S.~Legg, and D.~Amodei, ``Deep
  reinforcement learning from human preferences,'' in \emph{Advances in Neural
  Information Processing Systems (NeurIPS)}, vol.~30, 2017.

\bibitem{hejna2023few}
D.~J.~H. III and D.~Sadigh, ``Few-shot preference learning for
  human-in-the-loop rl,'' in \emph{Conference on Robot Learning (CoRL)}.\hskip
  1em plus 0.5em minus 0.4em\relax PMLR, 2023, pp. 2014--2025.

\bibitem{ji2023beavertails}
J.~Ji, M.~Liu, J.~Dai, X.~Pan, C.~Zhang, C.~Bian, B.~Chen, R.~Sun, Y.~Wang, and
  Y.~Yang, ``Beavertails: Towards improved safety alignment of llm via a
  human-preference dataset,'' in \emph{Advances in Neural Information
  Processing Systems (NeurIPS)}, vol.~36, 2023, pp. 24\,678--24\,704.

\bibitem{kirstain2023pick}
Y.~Kirstain, A.~Polyak, U.~Singer, S.~Matiana, J.~Penna, and O.~Levy,
  ``Pick-a-pic: An open dataset of user preferences for text-to-image
  generation,'' in \emph{Advances in Neural Information Processing Systems
  (NeurIPS)}, vol.~36, 2023, pp. 36\,652--36\,663.

\bibitem{wu2023human}
X.~Wu, K.~Sun, F.~Zhu, R.~Zhao, and H.~Li, ``Human preference score: Better
  aligning text-to-image models with human preference,'' in \emph{International
  Conference on Computer Vision (ICCV)}, October 2023, pp. 2096--2105.

\bibitem{bakker2022fine}
M.~Bakker, M.~Chadwick, H.~Sheahan, M.~Tessler, L.~Campbell-Gillingham,
  J.~Balaguer, N.~McAleese, A.~Glaese, J.~Aslanides, M.~Botvinick, and
  C.~Summerfield, ``Fine-tuning language models to find agreement among humans
  with diverse preferences,'' in \emph{Advances in Neural Information
  Processing Systems (NeurIPS)}, vol.~35, 2022, pp. 38\,176--38\,189.

\bibitem{wei2022chain}
J.~Wei, X.~Wang, D.~Schuurmans, M.~Bosma, b.~ichter, F.~Xia, E.~Chi, Q.~V. Le,
  and D.~Zhou, ``Chain-of-thought prompting elicits reasoning in large language
  models,'' in \emph{Advances in Neural Information Processing Systems
  (NeurIPS)}, vol.~35, 2022, pp. 24\,824--24\,837.

\bibitem{yao2023tree}
S.~Yao, D.~Yu, J.~Zhao, I.~Shafran, T.~Griffiths, Y.~Cao, and K.~Narasimhan,
  ``Tree of thoughts: Deliberate problem solving with large language models,''
  in \emph{Advances in Neural Information Processing Systems (NeurIPS)},
  vol.~36, 2023, pp. 11\,809--11\,822.

\bibitem{dantam2018incremental}
N.~T. Dantam, Z.~K. Kingston, S.~Chaudhuri, and L.~E. Kavraki, ``An incremental
  constraint-based framework for task and motion planning,''
  \emph{International Journal of Robotics Research (IJRR)}, vol.~37, no.~10,
  pp. 1134--1151, 2018.

\bibitem{marcucci2023motion}
T.~Marcucci, M.~Petersen, D.~von Wrangel, and R.~Tedrake, ``Motion planning
  around obstacles with convex optimization,'' \emph{Science Robotics}, vol.~8,
  no.~84, p. eadf7843, 2023.

\bibitem{mcdermott20001998}
D.~M. McDermott, ``The 1998 ai planning systems competition,'' \emph{AI
  Magazine}, vol.~21, no.~2, p.~35, 2000.

\bibitem{fox2003pddl2}
M.~Fox and D.~Long, ``Pddl2. 1: An extension to pddl for expressing temporal
  planning domains,'' \emph{Journal of Artificial Intelligence Research},
  vol.~20, pp. 61--124, 2003.

\bibitem{reda2020learning}
D.~Reda, T.~Tao, and M.~van~de Panne, ``Learning to locomote: Understanding how
  environment design matters for deep reinforcement learning,'' in
  \emph{Proceedings of the 13th ACM SIGGRAPH Conference on Motion, Interaction
  and Games}, 2020.

\bibitem{jorge2008planning}
J.~A. Baier and S.~A. McIlraith, ``Planning with preferences,'' \emph{AI
  Magazine}, vol.~29, no.~4, pp. 25--36, 2008.

\bibitem{canal2019adapting}
G.~Canal, G.~Aleny{\`a}, and C.~Torras, ``Adapting robot task planning to user
  preferences: an assistive shoe dressing example,'' \emph{Autonomous Robots},
  vol.~43, no.~6, pp. 1343--1356, 2019.

\bibitem{kim2017collaborative}
J.~Kim, C.~Banks, and J.~Shah, ``Collaborative planning with encoding of
  users’ high-level strategies,'' in \emph{AAAI Conference on Artificial
  Intelligence (AAAI)}, vol.~31, no.~1, 2017.

\bibitem{yu2023language}
W.~Yu, N.~Gileadi, C.~Fu, S.~Kirmani, K.-H. Lee, M.~G. Arenas, H.-T.~L. Chiang,
  T.~Erez, L.~Hasenclever, J.~Humplik, B.~Ichter, T.~Xiao, P.~Xu, A.~Zeng,
  T.~Zhang, N.~Heess, D.~Sadigh, J.~Tan, Y.~Tassa, and F.~Xia, ``Language to
  rewards for robotic skill synthesis,'' in \emph{Conference on Robot Learning
  (CoRL)}.\hskip 1em plus 0.5em minus 0.4em\relax PMLR, 2023, pp. 374--404.

\bibitem{eschmann2021reward}
J.~Eschmann, ``Reward function design in reinforcement learning,''
  \emph{Reinforcement learning algorithms: Analysis and Applications}, pp.
  25--33, 2021.

\bibitem{dong2022survey}
Q.~Dong, L.~Li, D.~Dai, C.~Zheng, J.~Ma, R.~Li, H.~Xia, J.~Xu, Z.~Wu, T.~Liu,
  \emph{et~al.}, ``A survey on in-context learning,'' \emph{arXiv preprint
  arXiv:2301.00234}, 2022.

\bibitem{min2022rethinking}
S.~Min, X.~Lyu, A.~Holtzman, M.~Artetxe, M.~Lewis, H.~Hajishirzi, and
  L.~Zettlemoyer, ``Rethinking the role of demonstrations: What makes
  in-context learning work?'' in \emph{Annual Conference on Empirical Methods
  in Natural Language Processing (EMNLP)}, 2022.

\bibitem{jiao2022sequential}
Z.~Jiao, Y.~Niu, Z.~Zhang, S.-C. Zhu, Y.~Zhu, and H.~Liu, ``Sequential
  manipulation planning on scene graph,'' in \emph{International Conference on
  Intelligent Robots and Systems (IROS)}, 2022, pp. 8203--8210.

\bibitem{rana2023sayplan}
K.~Rana, J.~Haviland, S.~Garg, J.~Abou-Chakra, I.~Reid, and N.~Suenderhauf,
  ``Sayplan: Grounding large language models using 3d scene graphs for scalable
  robot task planning,'' in \emph{Conference on Robot Learning (CoRL)}.\hskip
  1em plus 0.5em minus 0.4em\relax PMLR, 2023, pp. 23--72.

\bibitem{singh2023progprompt}
I.~Singh, V.~Blukis, A.~Mousavian, A.~Goyal, D.~Xu, J.~Tremblay, D.~Fox,
  J.~Thomason, and A.~Garg, ``Progprompt: Generating situated robot task plans
  using large language models,'' in \emph{International Conference on Robotics
  and Automation (ICRA)}, 2023, pp. 11\,523--11\,530.

\bibitem{zhu2021hierarchical}
Y.~Zhu, J.~Tremblay, S.~Birchfield, and Y.~Zhu, ``Hierarchical planning for
  long-horizon manipulation with geometric and symbolic scene graphs,'' in
  \emph{International Conference on Robotics and Automation (ICRA)}, 2021, pp.
  6541--6548.

\bibitem{armeni20193d}
I.~Armeni, Z.-Y. He, J.~Gwak, A.~R. Zamir, M.~Fischer, J.~Malik, and
  S.~Savarese, ``3d scene graph: A structure for unified semantics, 3d space,
  and camera,'' in \emph{International Conference on Computer Vision (ICCV)},
  October 2019.

\bibitem{chang2021comprehensive}
X.~Chang, P.~Ren, P.~Xu, Z.~Li, X.~Chen, and A.~Hauptmann, ``A comprehensive
  survey of scene graphs: Generation and application,'' \emph{Transactions on
  Pattern Analysis and Machine Intelligence (TPAMI)}, vol.~45, no.~1, pp.
  1--26, 2023.

\bibitem{yang2018graph}
J.~Yang, J.~Lu, S.~Lee, D.~Batra, and D.~Parikh, ``Graph r-cnn for scene graph
  generation,'' in \emph{European Conference on Computer Vision (ECCV)},
  September 2018.

\bibitem{agia2022taskography}
C.~Agia, K.~M. Jatavallabhula, M.~Khodeir, O.~Miksik, V.~Vineet, M.~Mukadam,
  L.~Paull, and F.~Shkurti, ``Taskography: Evaluating robot task planning over
  large 3d scene graphs,'' in \emph{Conference on Robot Learning (CoRL)}.\hskip
  1em plus 0.5em minus 0.4em\relax PMLR, 2022, pp. 46--58.

\bibitem{li2023behavior}
C.~Li, R.~Zhang, J.~Wong, C.~Gokmen, S.~Srivastava, R.~Mart{\'\i}n-Mart{\'\i}n,
  C.~Wang, G.~Levine, M.~Lingelbach, J.~Sun, \emph{et~al.}, ``Behavior-1k: A
  benchmark for embodied ai with 1,000 everyday activities and realistic
  simulation,'' in \emph{Conference on Robot Learning (CoRL)}.\hskip 1em plus
  0.5em minus 0.4em\relax PMLR, 2023, pp. 80--93.

\bibitem{coumans2021}
E.~Coumans and Y.~Bai, ``Pybullet, a python module for physics simulation for
  games, robotics and machine learning,'' \url{http://pybullet.org},
  2016--2021.

\bibitem{reimers-2019-sentence-bert}
N.~Reimers and I.~Gurevych, ``Sentence-bert: Sentence embeddings using siamese
  bert-networks,'' in \emph{Annual Conference on Empirical Methods in Natural
  Language Processing (EMNLP)}, 2019.

\bibitem{ding2023task}
Y.~Ding, X.~Zhang, C.~Paxton, and S.~Zhang, ``Task and motion planning with
  large language models for object rearrangement,'' in \emph{International
  Conference on Intelligent Robots and Systems (IROS)}, 2023, pp. 2086--2092.

\end{thebibliography}

\end{document}